\documentclass[conference]{IEEEtran}
\IEEEoverridecommandlockouts
\usepackage{cite}
\usepackage{amsmath,amssymb,amsfonts}
\usepackage{algorithmic}
\usepackage{graphicx}
\usepackage{textcomp}
\usepackage{xcolor}
\usepackage[a4paper, total={184mm,239mm}]{geometry}
\usepackage{hyperref}
\def\BibTeX{{\rm B\kern-.05em{\sc i\kern-.025em b}\kern-.08em
    T\kern-.1667em\lower.7ex\hbox{E}\kern-.125emX}}
\begin{document}

%\title{STEP-LLM: Towards Direct Generation of STEP Files for CAD Design with Large Language Models\\

\title{STEP-LLM: Generating CAD STEP Models from Natural Language with Large Language Models}

% \author{
%     Xiangyu Shi\textsuperscript{*}, Junyang Ding, Xu Zhao, Sinong Zhan, Payal Mohapatra, Daniel Quispe, \\ Kojo Welbeck, Jian Cao, Wei Chen, Ping Guo and Qi Zhu\textsuperscript{*} \\
%     Northwestern University
% }
\author{
\IEEEauthorblockN{
% Xiangyu Shi$^{1}$,
% Junyang Ding$^{2}$,
% Xu Zhao$^{1}$,
% Sinong Zhan$^{1}$,
% Payal Mohapatra$^{1}$,
% Daniel Quispe$^{3}$,
% Kojo Welbeck$^{3}$, \\
% Jian Cao$^{3}$,
% Wei Chen$^{3}$,
% Ping Guo$^{3}$,
% Qi Zhu$^{1}$ %
Xiangyu Shi,
Junyang Ding,
Xu Zhao,
Sinong Zhan,
Payal Mohapatra,
Daniel Quispe,
Kojo Welbeck, \\
Jian Cao,
Wei Chen,
Ping Guo,
Qi Zhu%
\thanks{Source code is available at \url{https://github.com/JasonShiii/STEP-LLM}. Corresponding authors: Xiangyu Shi (\textsf{xiangyushi2029@u.northwestern.edu}), Qi Zhu (\textsf{qzhu@northwestern.edu}). The work was supported in part by the National Science Foundation grants 2328032, 2324936, and 2328973.}
} 
\IEEEauthorblockA{
% $^{1}$Department of Electrical and Computer Engineering, Northwestern University\\
% $^{2}$Department of Computer Science, Northwestern University\\
% $^{3}$Department of Mechanical Engineering, Northwestern University
Northwestern University, Evanston, USA
%\\\texttt{xiangyushi2029@u.northwestern.edu, qzhu@northwestern.edu}
}
}

%\and
%\and
%\and
%\and
%\and

\maketitle

% \renewcommand{\thefootnote}{\fnsymbol{footnote}} % Optional: Use symbols like * instead of numbers
% \footnotetext[1]{Corresponding authors: \texttt{XiangyuShi2029@u.northwestern.edu}, \texttt{qzhu@northwestern.edu}}
% \footnote{Source code is available at \url{https://github.com/JasonShiii/STEP-LLM}}

\begin{abstract}
% Computer-aided design (CAD) is central to modern manufacturing, yet creating CAD models remains a specialized, labor-intensive process requiring significant expertise. This barrier limits rapid prototyping by designers without deep manufacturing knowledge. Recent works have applied large language models (LLMs) to text-to-CAD generation, typically targeting command sequences or script-based representations such as CadQuery. While useful, these formats are kernel-specific and lack the generality needed for downstream manufacturing. 
% In contrast, the STEP (ISO 10303) format is a widely adopted, neutral boundary representation (B-rep) standard that directly interfaces with manufacturing. Enabling LLMs to generate STEP models directly from natural language would empower non-experts to translate intuitive design intent into manufacturable artifacts. However, STEP is intrinsically graph-structured, with cross-references and non-sequential dependencies that challenge auto-regressive LLMs.
Computer-aided design (CAD) is vital to modern manufacturing, yet model creation remains labor-intensive and expertise-heavy.
%, hindering rapid prototyping. 
To enable non-experts to translate intuitive design intent into manufacturable artifacts, recent large language models (LLM)-based text-to-CAD efforts focus on command sequences or script-based formats like CadQuery. However, these formats are kernel-dependent and lack universality for manufacturing. In contrast, the Standard for the Exchange of Product Data (STEP, ISO 10303) file is a widely adopted, neutral boundary representation (B-rep) format directly compatible with manufacturing, but its graph-structured, cross-referenced nature poses unique challenges for auto-regressive LLMs.
% \simon{background intro is too detailed, one sentence as intro, and one more for motivation with literature}
To address this, we curate a dataset of $\sim$40K STEP-caption pairs and introduce novel preprocessing tailored for the graph-structured format of STEP, including a depth-first search (DFS)-based reserialization that linearizes cross-references while preserving locality and chain-of-thought(CoT)-style structural annotations that explicitly guide global coherence. We integrate retrieval-augmented generation (RAG) to ground predictions in relevant examples for supervised fine-tuning (SFT), and further refine generation quality through reinforcement learning (RL) with a specific Chamfer Distance-based geometric reward. Experiments demonstrate consistent gains of our STEP-LLM in geometric fidelity over the Text2CAD baseline, with improvements arising from multiple stages of our framework: the RAG module substantially enhances completeness and renderability, the DFS-based reserialization strategy strengthens overall accuracy, and the RL refinement further reduces geometric discrepancy. Both metrics and visual comparisons confirm that STEP-LLM generates shapes with higher fidelity than Text2CAD. These results demonstrate the feasibility of LLM-driven STEP model generation from natural language, showing its potential to democratize CAD design for manufacturing.
\end{abstract}


\section{Introduction}
\begin{figure}[!t]
  \centerline{\includegraphics[width=\linewidth]{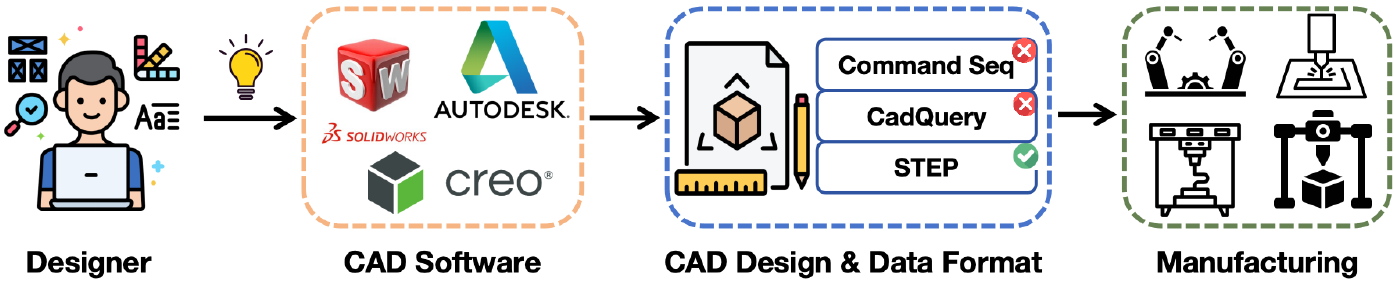}}
  \caption{The typical workflow of CAD from design to manufacturing.}
  \label{teaser}
  \vspace{-18pt}
\end{figure}

Computer-aided design (CAD) plays a foundational role in modern design and manufacturing. However, creating CAD models remains a specialized and labor-intensive process that requires extensive expertise and time, limiting accessibility for non-expert users and slowing the pace of prototyping. More broadly, the entire pipeline from design to manufacturing still relies heavily on manual intervention, with limited automation in translating design intent into directly manufacture models (Fig.~\ref{teaser}). Recently, the rapid progress of Generative AI has opened new possibilities for CAD workflow, raising the prospect that intuitive design intent expressed in natural language could be transformed into manufacture artifacts. Realizing this vision would allow non-experts to translate high-level ideas into manufacturable artifacts, lowering barriers for CAD design and manufacturing.
%prototype, thereby lowering barriers for fast prototyping.

Recent attempts at text-to-CAD generation have largely relied on \textbf{command sequences}~\cite{text2cad} or \textbf{script-based formats} such as CadQuery \cite{text-to-cadquery}. Command sequences represent the design history as a series of modeling operations (e.g., sketch, extrude) that can be replayed by a CAD kernel, while script-based formats express CAD models as executable code (e.g., Python or macros) that calls modeling APIs. These representations align well with large language models (LLMs)’ code generation abilities and allow validation through external engines\cite{zhang2025shop}. However, tied to specific kernels, these formats cannot directly support manufacturing and are constrained to simple objects by omitting complex operations such as fillets or free-form surfaces~\cite{brepgen}. In contrast, \textbf{boundary representation (B-rep)} encodes complete topology and geometry, providing the expressiveness needed to model complex designs. Recent works such as SolidGen~\cite{solidgen} and BrepGen~\cite{brepgen} demonstrate its learnability, yet remain focused on geometric validity rather than manufacturability. Within this family, the Standard for the Exchange of Product Data (STEP) file is a widely adopted, an engine-agnostic standard that \textbf{directly interfaces with industrial pipelines}, ensuring models can be used in downstream CAD/CAM. Although a few studies have analyzed STEP for machining feature recognition or entity parsing~\cite{apprach-step-as-language, ren4step}, direct generation of STEP files from natural language has not been explored. This gap motivates our work: leveraging LLMs to directly produce STEP models from natural language, combining LLMs’ generative strengths with STEP’s universality for manufacture-ready design.

However, direct STEP generation presents unique challenges. A STEP file is intrinsically \textbf{graph-structured}, with \textbf{cross-references and non-sequential dependencies} that conflict with the left-to-right auto-regressive paradigm of LLMs~\cite{Graph-KV}. In addition, small errors in entity ordering or identifier usage can render the entire file invalid. 

% Overcoming this challenge requires both novel data preprocessing and tailored training strategies.

To address this, we propose STEP-LLM, a complete fine tuning framework for natural language to STEP generation based on state-of-the-art small models. We first construct a dataset consisting around 40K caption-STEP pairs, and introduce a depth-first search (DFS)-style reserialization strategy that linearizes STEP’s graph structure to preserve \textbf{local sequential ordering}. Moreover, CoT-style annotations are used to guide \textbf{global coherence}. We further integrate retrieval-augmented generation (RAG) into supervised fine-tuning (SFT), grounding outputs in relevant examples. In addition, we generalize model's reasoning capability through reinforcement learning (RL) with a specific scaled Chamfer Distance–based geometric reward, which is robust to translation, orientation, and scale differences, providing a reliable objective for optimizing shape fidelity.

In the absence of broadly established benchmarks, we evaluate our method against the Text2CAD~\cite{text2cad} baseline on renderability, geometric fidelity, and the alignment of generated STEP file complexity with the distribution of ground-truth models. In addition, we conduct a comprehensive ablation study to verify the contribution of each design choice, including RAG, DFS-based reserialization, RL, and base model selection. Experiment results demonstrate that our framework achieves high completion and renderability rates, while significantly improving geometric accuracy compared to baselines. Our main contributions are as follows:
\begin{itemize}
    \item We propose \textbf{STEP-LLM}, the first unified framework for direct STEP file generation from natural language, bridging LLMs with a universal CAD standard. The framework integrates RAG to enrich training context and augment SFT, and RL with geometry-aware rewards to further enhance geometric fidelity and robustness.  
    \item We introduce a \textbf{novel DFS-based reserialization strategy} for STEP file preprocessing, which linearizes the graph-like structure of STEP files to better align with the auto-regressive nature of LLMs. In addition, we incorporate \textbf{CoT-style structural annotations} to guide global coherence, enabling the model to capture both local sequential ordering and long-range dependencies.  
    \item We release a \textbf{curated dataset} of caption–STEP pairs and a set of \textbf{evaluation metrics} for benchmarking text-to-STEP generation, facilitating future research in this area.  
\end{itemize}

\section{Background and Related Work}
% Recent works have begun exploring the use of LLMs for CAD generation, with most efforts focusing on command sequences \cite{cad-gpt, text2cad, flexcad, cad-mllm, cad-llama, img2cad, cadvlm, openecad} or script-based formats such as CadQuery \cite{text-to-cadquery, cad-coder-image, cad-coder-text, LLM4CAD} and CAD Macros \cite{query2cad}. In parallel, there has been progress in learning-based methods for Boundary Representation (B-rep) modeling and analysis \cite{brepgen, polygen, complexgen, solidgen, point2cad}. However, to date, no approach has leveraged LLMs to directly generate B-rep representations.
\subsection{Large Language Models for CAD Generation}
Recent works have explored applying LLMs to CAD generation, primarily focusing on command sequences~\cite{cad-gpt, text2cad, flexcad, cad-mllm, cad-llama, img2cad, cadvlm, openecad} or script-based formats such as CadQuery~\cite{text-to-cadquery, cad-coder-image, cad-coder-text, LLM4CAD} and CAD Macros~\cite{query2cad}. Command sequences, popularized by datasets such as DeepCAD~\cite{DeepCAD}, represent CAD design history as procedural instructions recording the modeling operations. FlexCAD~\cite{flexcad} converts a CAD model into a brief, structured text and employs hierarchy-aware masking to fine-tune LLMs for controllable CAD generation tasks. CAD-Llama~\cite{cad-llama} follows an adaptive pretraining paradigm combined with instruction tuning on a multitask instructional dataset for multi-task modeling. Several works~\cite{cad-gpt, cad-mllm} further expand the input to multi-modality such as image, point cloud. Script-based formats express CAD models as executable code or scripts that call modeling APIs in a human-readable style. Representative examples include Query2CAD~\cite{query2cad}, which uses FreeCAD macros with self-refinement loops, and CAD-Coder~\cite{cad-coder-text}, which translates natural language into CadQuery programs. Due to LLMs’ strong capability in code generation, these approaches show clear advantages in producing syntactically valid and easily verifiable CAD scripts.

Despite these advances, both command-sequence and script-based approaches rely heavily on external CAD engines for execution, limiting universality across design platforms. Moreover, the expressiveness of these formats is constrained by the APIs of the target engine or the vocabulary of the sequence, restricting their ability to capture complex operations such as fillet, chamfer, or B-spline surfaces. Therefore, while effective for controlled tasks, these representations are not directly suitable for downstream manufacturing, motivating the need for more universal CAD formats.

\subsection{Learning-based B-rep Generation and STEP File Analysis}
Unlike command sequences or script-based formats, which abstract away from geometric detail, B-rep encodes the complete topology and geometry of a solid model, including vertices, edges, faces, and their connectivity, making it inherently capable of describing complex operations. Several works in geometric deep learning addressed B-rep generation directly. SolidGen~\cite{solidgen} employed an auto-regressive strategy to synthesize vertices, edges, and faces in sequence. BrepGen~\cite{brepgen} advanced this line with a diffusion-based model over a structured latent geometry tree, enabling generation of free-form surfaces. ComplexGen~\cite{complexgen} proposed a chain-complex view of B-reps, reconstructing corners, curves, and surface patches jointly, while Point2CAD~\cite{point2cad} extended this paradigm to reverse-engineer B-reps from point clouds. These studies confirm that the graph-structured nature of B-reps is learnable and expressive, yet they remain primarily theoretical explorations focused on geometric validity rather than manufacturability and cannot be directly deployed in industrial workflows.

Along with these advances, the STEP file provides a standardized textual realization of B-rep and has become one of the most universal, engine-independent formats for exchanging product manufacturing information. A STEP file is organized into two main sections: the HEADER SECTION, which specifies metadata such as schema, author, and units, and the DATA SECTION, which encodes the actual geometry and topology. In the DATA SECTION, models are represented as a collection of entities (e.g., CARTESIAN\_POINT, EDGE, FACE) linked together by unique identifiers, enabling cross-references that capture the full structure of the design. Limited prior work has treated STEP file as a structured language~\cite{apprach-step-as-language}, using recursive neural networks for tasks such as machining feature recognition and entity analysis~\cite{ren4step}. These efforts demonstrate the feasibility of parsing STEP files directly, but to the best of our knowledge, none has attempted LLM-based generation of STEP. This gap motivates our work: we investigate whether an LLM can be taught to “speak” the language of STEP, combining the generative strengths of modern language models with the universality and manufacturability of a CAD standard.

\begin{figure*}[!t]
  \centering
  \includegraphics[width=\textwidth]{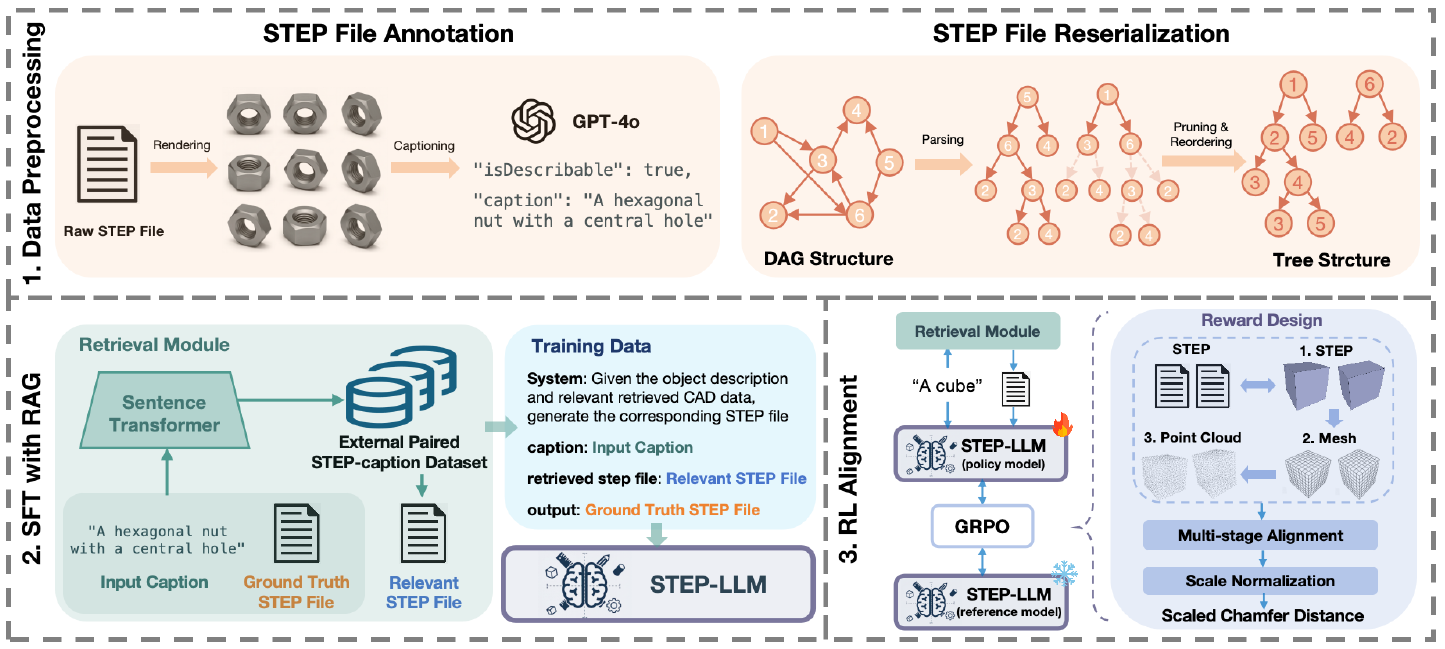}
  \caption{Framework of STEP-LLM. The top panel illustrates data preprocessing: raw STEP files are rendered into nine views and captioned with GPT-4o (left), and their internal DAG structures (the numbers represent the identifier of the entity in a STEP file) are reserialized into locality-preserving trees (right). These captions and DFS-style STEP files together form the paired STEP-caption dataset for SFT (bottom-left). A specific geometric reward based on scaled Chamfer Distance is designed for further RL training (bottom-right).}
  \label{framework}
    \vspace{-12pt}
\end{figure*}

\section{The STEP-LLM Framework}

% To tackle the challenge of direct STEP generation from natural language, our framework (Figure~\ref{framework}) consists of three main components. First, we curate and preprocess a paired dataset of captions and STEP files (Section~\ref{DP}), which involves rendering multi-view images for captioning and reserializing STEP files into locality-preserving sequences with structural annotations. Second, we perform supervised fine-tuning (SFT) with retrieval-augmented generation (RAG) (Section~\ref{SFT}), allowing the model to learn STEP grammar and leverage relevant prior examples for more consistent outputs. Finally, we refine the model with reinforcement learning alignment (Section~\ref{RL}), where a geometric reward based on scaled Chamfer Distance (SCD) further encourages syntactic validity and geometric fidelity. Together, these stages form a coherent pipeline that grounds training in structured data, augments it with retrieval, and aligns it with task-specific evaluative criteria.
Our framework (Fig.~\ref{framework}) consists of three main stages. First, we construct a paired caption–STEP dataset through rendering, captioning, and reserialization (Section~\ref{DP}). Second, we perform supervised fine-tuning (SFT) with RAG (Section~\ref{SFT}), allowing the model to learn STEP grammar and leverage relevant prior examples for more consistent outputs. Finally, we refine the model with RL alignment (Section~\ref{RL}), where a geometric reward based on scaled Chamfer Distance further improves geometric fidelity.

\subsection{Dataset Construction and Preprocessing}\label{DP}
% We build our dataset upon the ABC dataset \cite{ABC}, a large-scale collection of CAD models released for geometric deep learning. The ABC dataset contains over one million B-rep models exported to STEP format, alongside corresponding parametric curves and surface annotations.

% To control training efficiency and account for model complexity, we adopt the entity number of a STEP file as an estimation for shape complexity. Two subsets are constructed: (i) all models containing fewer than 500 entities are used as the main training set for SFT, resulting in $\sim$20,500 models, and (ii) models with 500-1000 entities are reserved for continuous SFT in our scalability experiments (Section IV), yielding $\sim$17,000 models. This stratification enables us to systematically evaluate how model performance scales with increasingly complex CAD geometries. 
We build our dataset upon the ABC dataset~\cite{ABC}, a large-scale collection of CAD models released for geometric deep learning. The ABC dataset contains over one million B-rep models exported to STEP format, alongside corresponding parametric curves and surface annotations. To control training efficiency and account for model complexity, we adopt the entity number of a STEP file as a proxy for shape complexity, preparing a filtered dataset for proceeding supervised fine-tuning (see Section~\ref{SFT}). This selection strategy ensures that the data scale and difficulty are appropriate for LLM training.

% To construct text-STEP paired data suitable for LLM training, we first generate a nine-view rendered image (resolution 1200×1200) for each STEP file using standard orthographic and perspective projections. We then employ GPT-4o to first determine whether these images can be described, and then caption those describable images. Notably, we observe that current LLMs, including GPT-4o, exhibit notable limitations when tasked with describing multiview renderings: they often struggle with \textbf{perspective consistency} and may incorrectly interpret multiple views of the same object as distinct items.

To construct text–STEP paired data, we render each STEP file into multi-view images (9 views, resolution 1200×1200) following the common practice in prior work such as Text2CAD, and as illustrated in our system framework (Fig.~\ref{framework}). These include both orthographic and perspective views, providing comprehensive geometric coverage. We then employ GPT-4o to assess the renderings and generate captions for those deemed describable. In practice, however, we observe that current foundation models, including GPT-4o, face challenges in describing multi-view renderings: they often lack \textbf{perspective consistency} and may incorrectly interpret different views of the same object as multiple objects\cite{wang2023empowering,li2025shedding, zhan2025sentinel}.

To overcome this limitation, we apply tailored prompt engineering to guide GPT-4o’s captioning process. Our prompts encourage concise yet informative textual descriptions that emphasize salient geometric features (e.g., symmetry, through holes, fillets) while also linking objects to real-world categories whenever appropriate. For example, a model might be captioned as \textit{“A flat circular lid with two rectangular mounting tabs and a central recessed feature.”} This strategy produces captions that are human-interpretable and sufficiently structured to supervise STEP file generation in subsequent model training.

% \subsection{ STEP File Reserialization with CoT annotation}

% The STEP file format, formally standardized as ISO 10303, is one of the most widely adopted neutral representations for CAD models. Each STEP file encodes a 3D shape as a collection of entities (e.g., EDGE, FACE, CARTESIAN\_POINT), with dependencies expressed through cross-referencing identifiers. From a structural perspective, a STEP file can be regarded as a \textbf{directed acyclic graph (DAG)} or a \textbf{pointer-based graph}, in which high-level shape definitions recursively reference lower-level primitives. This graph-based nature makes STEP universal and expressive, but it also introduces challenges when used as training targets for LLMs.

% In its raw textual form, the reference relationship of entities within a STEP file is highly mixed and non-sequential: entities that depend on one another may be separated by tens or even hundreds of lines, requiring long-range memorization to maintain reference consistency. For a transformer-based LLM, which generates outputs in a strictly left-to-right auto-regressive paradigm, this poses a significant difficulty. The model must remember and correctly reuse identifiers created far earlier in the sequence, and small errors in entity ordering or referencing can easily lead to unrecoverable invalid outputs.

From a structural perspective, STEP files are challenging targets for LLMs. Each file encodes a model as a collection of entities connected by cross-references, forming a \textbf{directed acyclic graph (DAG)} that captures the hierarchical relationships between primitives and higher-level geometry. However, in their raw textual form this DAG structure becomes highly non-sequential: related entities may be scattered far apart, and long-range identifier dependencies must be recalled precisely. For an auto-regressive model, this easily leads to incoherent or invalid outputs.

To mitigate this issue, we introduce a DFS-based reserialization strategy. Specifically, we first parse the STEP file into a \textbf{hierarchical tree structure} where each node represents an entity and its child nodes correspond to directly referenced entities. We then serialize the tree using a depth-first traversal. This approach allows each branch of the tree to be expressed as a relatively local and coherent sequence, reducing the burden of long-range dependency tracking. To prevent entity explosion when traversing large graphs, we employ \textbf{strategic pruning}, ensuring only structurally relevant branches are expanded and each reference relationship appears once. Furthermore, we \textbf{renumber entity identifiers sequentially}, eliminating irregular gaps in the raw file and simplifying reference tracking. We also normalize floating-point precision, reducing unnecessary digits while preserving geometric validity, thereby lowering textual complexity without altering topology.

Nevertheless, DFS traversal alone cannot fully resolve discontinuities when switching between parallel branches. For example, once one branch of an entity is fully expanded, the model must switch context to a different branch of the same parent entity. Such discontinuities can cause the model to lose coherence. Inspired by recent progress in code generation with CoT-style annotations~\cite{survey-llm-code}, we augment the reserialized STEP files with lightweight statistical annotations. These annotations summarize branch-level statistics (e.g., number of child entities, branch depth) and act as guidance tokens, helping the LLM to reason about the \textbf{global structure} while maintaining consistency across branch transitions.

Overall, this DFS-based reserialization, combined with structural annotations, preserves locality while enhancing global comprehensibility. This strategy also mitigates the \textbf{inherent heterogeneity} of STEP files, where different entity orders may in fact correspond to the same model, thereby providing a more consistent representation for downstream learning.

% \begin{figure}[htbp]
% \centerline{\includegraphics[width=\linewidth]{reserialization.png}}
% \caption{The comparison between raw format and reserialized dfs-like format(indented for clarity)}
% \label{raw-dfs}
% \end{figure}

\subsection{Supervised Fine-Tuning with RAG} \label{SFT}

Supervised fine-tuning is an essential step in adapting LLMs for domain-specific applications~\cite{SFT4domain,zhang2025shop, mohapatra-etal-2025-llms}. In our setting, SFT allows the model to internalize the grammar of STEP files and learn the mapping between natural language captions and precise geometric representations.

To enhance this process, we integrate RAG into the fine-tuning pipeline, which enriches the model’s context by grounding generation in semantically relevant prior examples. This approach reduces the reliance on parameter memorization and promotes output faithfulness. Such grounding is particularly critical for STEP generation, where long-range dependencies and structural consistency challenges LLMs.

Our retrieval module is implemented as follows. We construct an external database of paired STEP files and captions. For a given input caption, the system retrieves the most semantically similar caption and its associated STEP file from the database. Specifically, we use SentenceTransformer~\cite{sentenceTransformer} to embed captions into dense vectors and FAISS~\cite{faiss} to index them for efficient nearest-neighbor search based on cosine similarity, ensuring that structurally relevant cases are retrieved. During SFT, the input prompt is augmented with both the original caption and the retrieved STEP file, while the target output remains the ground-truth STEP file of the original caption. This design provides the model with structural cues while still requiring faithful reproduction of the target object.

We adopt SFT on a curated subset of STEP files with fewer than 500 entities. Complex STEP files with higher entity counts (500–1000) often contain long sequences of repetitive entities, such as CARTESIAN\_POINT or DIRECTION, which can cause LLMs to fall into repetitive loops and fail to produce complete files. By focusing training on simpler files, the model effectively learns fundamental STEP syntax and compositional rules, which improves stability and robustness when generalizing to more complex cases. This strategy enhances both performance and reliability in generating valid STEP files.

\subsection{Reinforcement Learning Alignment} 
\label{RL}
Recent advances in natural language processing have shown that RL can further improve model performance beyond supervised fine-tuning~\cite{RL,feng2025token}. While SFT equips the model with the grammar of STEP files and the ability to map captions to geometry, it does not guarantee that generated outputs are geometrically faithful. In RL, one of the most critical factors is the \textbf{reward design}\cite{zhang2025lessons,zhang2025shop,zhan2024model}. To explicitly encode geometric fidelity into the training objective, we design a reward function based on the \textbf{Scaled Chamfer Distance (SCD)}, which is robust to translation, orientation, and scale differences. We implement this reward within a standard policy optimization framework using Group Relative Policy Optimization (GRPO)~\cite{GRPO}, and observe clear improvements in geometric accuracy.

\textbf{Scaled Chamfer Distance (SCD).} To evaluate geometric fidelity, we first compute the Chamfer Distance (CD) between the generated model and the ground truth. Both STEP files are converted into STL meshes, from which point clouds are sampled. Given point sets $P$ (prediction) and $Q$ (ground truth), the bidirectional Chamfer Distance is defined as:
\begin{equation}
\begin{split}
   &\text{CD}(P, Q) = \\
&\frac{1}{|P|}\sum_{p \in P}\min_{q \in Q}\|p - q\|_2^2 
+ \frac{1}{|Q|}\sum_{q \in Q}\min_{p \in P}\|q - p\|_2^2 \label{CD} 
\end{split}
\end{equation}
However, raw CD is sensitive to translation, orientation, and scale. To ensure fair comparison, we apply a multi-stage alignment strategy:
\begin{itemize}
\item \textbf{Center alignment:} shift both point clouds so that their centroids coincide, removing large translations.  
\item \textbf{Global registration:} compute a coarse rigid transform via feature matching (e.g., FPFH + RANSAC).
\item \textbf{Iterative Closest Point (ICP):} iteratively refine the alignment for precise point-to-point correspondence. 
\end{itemize}

We then normalize the distance by dividing it with the squared scale factor:
\begin{equation}
    \text{SCD} = \frac{\text{CD}}{(\text{Scale Factor})^2}, \label{scale_CD}
\end{equation}
where the scale factor is defined as the root mean square distance of ground-truth points from its centroid.

\textbf{Reward design.} Inspired by CAD-Coder~\cite{cad-coder-text}, we adopt a piecewise linear reward function with two thresholds. If the SCD is below the lower bound, the reward is 1; if above the upper bound, the reward is 0; for intermediate values, the reward is interpolated linearly to avoid sparse rewards:
\begin{equation}
R^{\text{geo}}(S) =
\begin{cases}
1, & \text{if } \text{SCD}(S, S_{\text{gt}}) \leq \delta_{\text{low}}, \\
0, & \text{if } \text{SCD}(S, S_{\text{gt}}) \geq \delta_{\text{high}}, \\
\frac{\delta_{\text{high}} - \text{CD}(S, S_{\text{gt}})}{\delta_{\text{high}} - \delta_{\text{low}}}, 
& \text{otherwise}.
\end{cases}
\label{CD_reward}
\end{equation}

This design encourages the model to generate STEP files that not only remain syntactically valid but also yield geometrically accurate shapes, with robustness to common variations in scale, position, and orientation.

\section{Experiments and Evaluation}

\subsection{Experimental Setup}
We conduct our SFT on two representative open-source models: Llama-3.2-3B-Instruct and Qwen-2.5-3B. All training is performed using the Unsloth~\cite{unsloth} framework for parameter-efficient fine-tuning. Our training corpus consists of 14,396 STEP-caption pairs, restricted to models with fewer than 500 entities. Additionally. we adopt LoRA~\cite{LoRA} to reduce GPU memory consumption and improve efficiency. The training configuration includes an effective batch size of 16, the AdamW (8-bit) optimizer, and a learning rate of 2e-4 with a linear decay scheduler. A warm-up of 5\% of the total training steps is applied to stabilize early training. The models are trained for 10 epochs on a single NVIDIA A100 GPU. To facilitate evaluation, we save a checkpoint after each epoch and perform asynchronous validation. 

For evaluation, we adopt a set of metrics that together capture syntactic validity, structural correctness, and geometric fidelity. Three primary metrics are used: Completion Rate (CR), Renderability Rate (RR), and Median Scaled Chamfer Distance (MSCD). CR verifies whether a generated STEP file terminates correctly, RR checks whether it can be reconstructed into a valid 3D object by OpenCASCADE~\cite{pyOCCT}, and MSCD quantifies geometric fidelity after alignment and scale normalization. Since baselines such as Text2CAD rely on command sequences and export STEP files through external engines, CR is not directly applicable in baseline comparisons. To address this, we additionally report Average Entity Count (AEC) as a proxy for design complexity, where a closer match between generated and ground-truth entity counts indicates stronger alignment with the expected difficulty level. More specifically, the metrics are as follows:
\begin{itemize}
    \item \textbf{Completion Rate (CR)}: percentage of generated STEP files that terminate correctly with the standardized "END-ISO-10303-21;" line.
    \item \textbf{Renderability Rate (RR)}: proportion of files successfully parsed by OpenCASCADE into non-null shapes that can be meshed into valid STL files.
    \item \textbf{Median Scaled Chamfer Distance (MSCD)}: median of scale-normalized Chamfer Distances across the test set, robust to translation, rotation, and scale differences.
    \item \textbf{Average Entity Count (AEC)}: mean number of entities in generated STEP files compared to ground truth. 
\end{itemize}

% Despite their relatively modest scale (3B parameters), both Llama-3.2 and Qwen-2.5 demonstrate encouraging performance in generating syntactically valid and geometrically faithful STEP files. These results suggest that STEP generation is feasible even with compact open-source LLMs, and provide a solid foundation for subsequent reinforcement learning.

\subsection{Advantages of STEP-LLM over Baseline}

\begin{table}[!b]
\centering
\caption{Comparison with Text2CAD baseline. RR: renderability rate; MSCD: median scaled Chamfer distance; AEC: avg. entity count (ground truth: 265.64).}
\label{tab:baseline}
\footnotesize
\setlength{\tabcolsep}{12pt}        % column padding
\renewcommand{\arraystretch}{1.05} % row height
\begin{tabular}{lccc}
\hline
\textbf{Method} & \textbf{RR (\%) $\uparrow$} & \textbf{MSCD $\downarrow$} & \textbf{AEC} \\
\hline
Text2CAD & 98.38 & 3.99 & 390.41 \\
STEP-LLM     & 95.18 & 0.53 & 240.99 \\
\hline
\end{tabular}
\end{table}

\begin{table}[t]
\centering
\footnotesize
\setlength{\tabcolsep}{12pt}
\caption{Entity count statistics on the test set.}
\label{tab:entity_stats}
\begin{tabular}{lccc}
\hline
\textbf{Method} & \textbf{Avg.} & \textbf{Min} & \textbf{Max} \\
\hline
Ground Truth & 265.64 & 47  & 477  \\
Text2CAD     & 390.41 & 50  & 3665 \\
STEP\textendash LLM   & 240.99 & 47  & 989  \\
\hline
\end{tabular}
\end{table}

\begin{figure}[!t]
\centerline{\includegraphics[width=\linewidth]{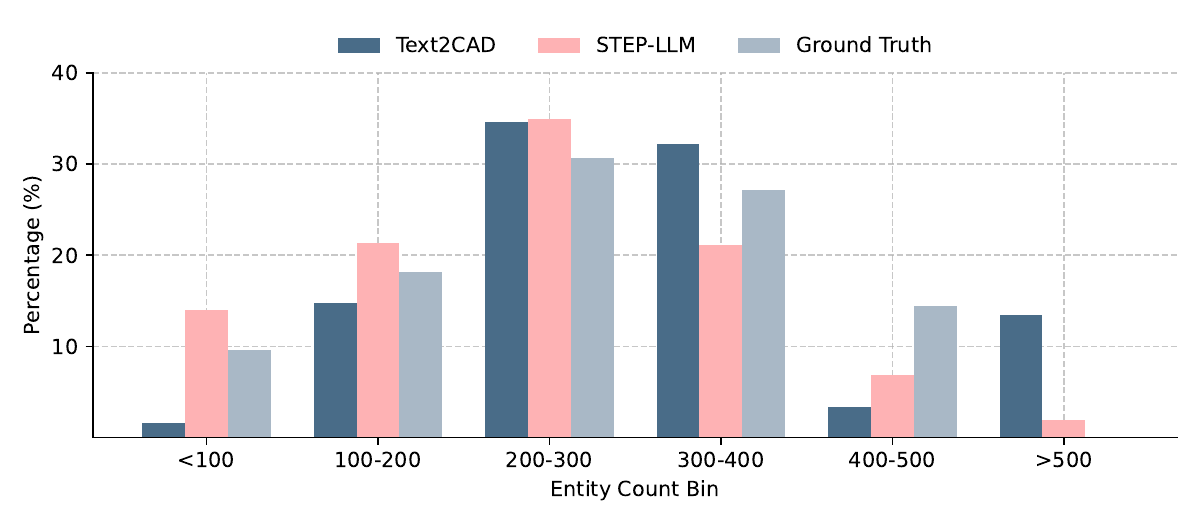}}
\caption{Entity count distribution of generated STEP files vs. ground truth.}
\label{entity_distribution}
\vspace{-12pt}
\end{figure}
Evaluating text-to-CAD generation remains challenging as the field is still in its early stage, and there are no widely adopted benchmarks or standardized baselines~\cite{text2cad, CAD-Judge}. Among the available open-sourced approaches, we select Text2CAD~\cite{text2cad} as the primary baseline. Published recently in NeurIPS 2024 as a spotlight paper, Text2CAD represents the strongest openly available method for text-to-parametric CAD generation, and has quickly become the standard point of comparison in this emerging area~\cite{cad-llama, CAD-Judge}. While several recent works report improvements under their own closed settings, their models or code are not publicly accessible, making Text2CAD the only suitable open-source baseline for fair and reproducible evaluation. In addition, it is capable of directly outputting STEP files during inference, which allows fair comparison and seamless integration into our evaluation pipeline without requiring additional kernel-dependent conversions. 

We evaluate both methods on a held-out set of 2,056 samples. The results are summarized in Table~\ref{tab:baseline}. \textbf{Our STEP-LLM delivers a renderability ratio comparable to Text2CAD, while achieving a substantially lower median scaled Chamfer distance, indicating higher geometric fidelity.} It is worth noting that Text2CAD benefits from relying on a CAD-kernel-based reconstruction and export pipeline, which enforces syntactic validity of STEP files and thus naturally yields a slightly higher renderability rate. In contrast, our framework generates STEP files directly, which makes the renderability criterion more stringent. Nevertheless, the obtained RR of over 95\% shows that our approach remains robust and practically reliable, especially considering our clear advantage in geometric fidelity. 

Beyond renderability and geometric distance, we further analyze the distribution of entity counts in generated STEP files (Table~\ref{tab:entity_stats} and Fig.~\ref{entity_distribution}). Our results show that \textbf{STEP-LLM produces entity counts whose average and distribution are much closer to the ground truth}, whereas Text2CAD often generates excessively large structures, with some files exceeding 3,000 entities. Such over-generation highlights instability in Text2CAD, while our method better preserves the realistic complexity range observed in human-designed CAD models.

\begin{figure}[!t]
\centerline{\includegraphics[width=\linewidth]{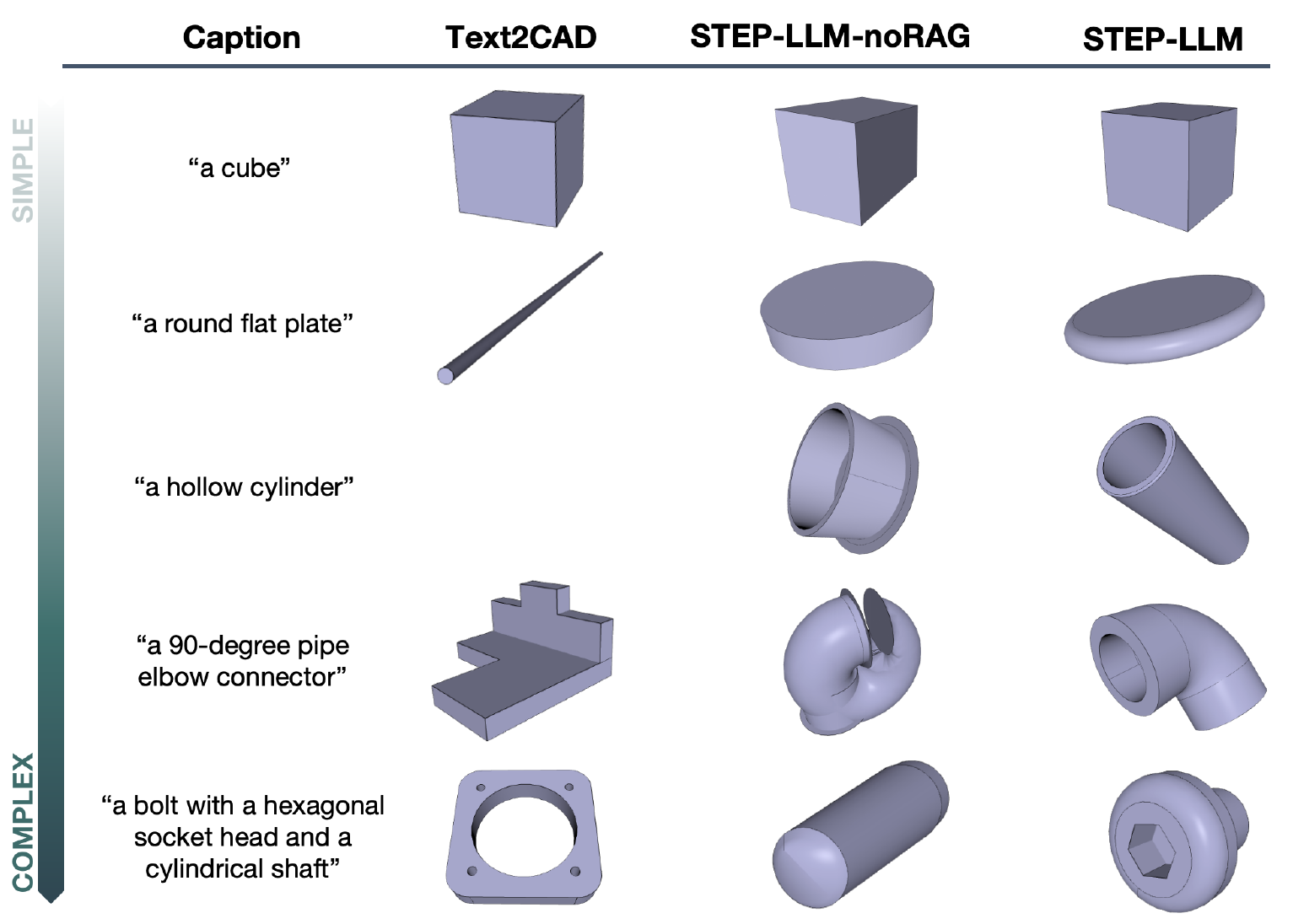}}
\caption{The visual comparison between Text2CAD, STEP-LLM-noRAG and STEP-LLM.}
\label{visual_compare}
\end{figure}

Finally, Fig.~\ref{visual_compare} shows qualitative results across different complexity levels. For both simple and complex prompts, STEP-LLM produces outputs that are visually more faithful to the described objects than Text2CAD, with notable improvements on curved geometries and detailed features. These results further demonstrate the effectiveness of our framework in advancing STEP file generation beyond current baseline.

\subsection{Ablation Studies on Module Importance}
To better understand the contributions of individual components in our framework, we conduct an ablation study comparing different configurations of model architecture and data processing. Specifically, we examine (i) the effect of DFS reserialization versus training on raw STEP files, (ii) the benefit of RAG versus non-RAG conditions, and (iii) differences across backbone models (Llama-3.2-3B-Instruct and Qwen-2.5-3B).

The results are summarized in Table~\ref{tab:ablation}. Our full configuration, which combines DFS-based reserialization and RAG with a Llama base model (denoted as Llama-dfs-RAG), achieves the best trade-off among completion rate, renderability rate, and geometric fidelity, and is therefore adopted as the backbone of STEP-LLM. Removing either dfs (Llama-nodfs-RAG) or RAG (Llama-dfs-noRAG) leads to significantly worse performance, \textbf{showing the importance of both modules in our approach}. Note that when comparing RAG with no-RAG conditions, to ensure fairness, we trained all models for the same number of steps. Since RAG prompts are longer, this means the RAG condition processed more tokens overall. We regard step-matching as the fairer comparison, as it keeps the optimizer schedule identical across conditions. Finally, Qwen backbone model provides similar performance as Llama, and we chose the latter due to its better results in geometric fidelity (MSCD).

\begin{table}[!t]
\centering
\caption{Ablation studies on STEP reserialization, RAG, and backbone models. Bold indicates best result per column.}
\label{tab:ablation}
\footnotesize
\begin{tabular}{lccc}
\hline
\textbf{Model Setting} & \textbf{CR (\%) $\uparrow$} & \textbf{RR (\%) $\uparrow$} & \textbf{MSCD $\downarrow$} \\
\hline
Llama-dfs-RAG      & 0.97 & \textbf{0.95} & \textbf{0.53} \\
Llama-nodfs-RAG      & 0.91 & 0.76          & 0.95 \\
Llama-dfs-noRAG    & 0.84 & 0.13         & 0.61 \\
Qwen-dfs-RAG       & \textbf{0.99} & 0.94 & 0.59 \\
\hline
\end{tabular}
\end{table}

% \subsection{Continuous SFT \& Scalability Analysis}
% Present results of your best model (DFS + RAG + RLHF) on the 1000 entity dataset. Analyze performance degradation and discuss challenges.

% Continuous SFT with rehersal: further finetuned on even small scale more complex data improves our performance: shows the potential and scalability with our framework. (check the #ave entity number of completed STEP file and compare with the ave gt output length to evaluate the matching level)

\subsection{Effectiveness of RL}
We further explore the effectiveness of RL for STEP file generation using GRPO. For the threshold in reward design, we set the upper bound to 0.5 and the lower bound to 0.01 to ensures that the model is encouraged toward geometrically accurate generations while avoiding sparse reward issues. We cold start RL training from the previous SFT checkpoint of Llama3.2-3B-Instruct, and conduct GRPO optimization with the following hyperparameters: batch size of 8, 8 sampled responses per prompt, KL penalty coefficient of 0.02, entropy coefficient of 0.005, and a learning rate of 3e-6. Training was conducted on 4 NVIDIA H100 GPUs for a total of 80 optimization steps.

The results are summarized in Table~\ref{tab:rl}. Compared to the supervised baseline, RL refinement leads to further improvement in completion ratio and a significant reduction in median Chamfer Distance, \textbf{demonstrating the effectiveness of even a small number of RL updates in enhancing geometric fidelity}. While the renderability rate shows a minor decline, this reflects the inherent trade-off introduced by optimizing for fine-grained geometric accuracy, as the model is encouraged to produce richer details. Importantly, the overall reliability remains high, confirming the effectiveness of using RL.

\begin{table}[!t]
\centering
\caption{Results of RL refinement on Llama3.2-3B-Instruct with Chamfer Distance reward.}
\label{tab:rl}
\footnotesize
\begin{tabular}{lccc}
\hline
\textbf{Model} & \textbf{CR (\%) $\uparrow$} & \textbf{RR (\%) $\uparrow$} & \textbf{MSCD $\downarrow$} \\
\hline
STEP-LLM & 0.97 & 0.95 & 0.53 \\
STEP-LLM-GRPO & \textbf{0.99} & 0.92 & \textbf{0.098} \\
\hline
\end{tabular}
\end{table}

\section*{Conclusion}

In this work, we introduced STEP-LLM, the first unified framework for direct LLM-based generation of STEP files. It incorporates four key components: DFS-based reserialization to linearize STEP’s graph structure, CoT-style annotation to ensure global coherence, retrieval-augmented fine-tuning to improve renderability, and reinforcement learning with geometry-aware rewards to refine fidelity and robustness. Together, these components enable the generation of syntactically valid, geometrically accurate, and manufacture-ready STEP representations.

%Due to the limited availability of open-source text-to-CAD models, our evaluation is restricted to comparison with a single baseline. 
Looking ahead in future work, scaling to larger models, refining reward functions with manufacturability or constraint checks, and expanding the dataset with richer captions and diverse CAD domains may further improve the model's performance.
% \noindent\textbf{Limitations.}Due to less available models, we use limited baseline comparison. We hope our.... 

\bibliographystyle{IEEEtran}
\bibliography{references}

% \vspace{12pt}
% \color{red}
% IEEE conference templates contain guidance text for composing and formatting conference papers. Please ensure that all template text is removed from your conference paper prior to submission to the conference. Failure to remove the template text from your paper may result in your paper not being published.

\end{document}